\documentclass[]{article}
\usepackage{authblk}
\usepackage{fullpage}

\usepackage[utf8]{inputenc} 
\usepackage{hyperref}       
\usepackage{url}            
\usepackage{booktabs}       
\usepackage{amsfonts}       
\usepackage{nicefrac}       
\usepackage{microtype}      
\usepackage{graphicx}
\usepackage{subfig}
\usepackage{xcolor}
\usepackage{multirow}
\usepackage{diagbox}
\usepackage{Definitions}
\newcommand{\xdownarrow}[1]{{\left\downarrow\vbox to #1{}\right.\kern-\nulldelimiterspace}}
\usepackage{algorithm,algorithmic}
\RequirePackage{amsmath,amssymb}
\RequirePackage{colortbl,slashbox}
\RequirePackage[disable,textsize=scriptsize,backgroundcolor=yellow!20]{todonotes}
\RequirePackage{enumitem} \setlist[description]{itemsep=.3ex}
\RequirePackage[skip=.3ex minus.1ex]{caption}
\definecolor{LightBlue}{rgb}{0.90,0.98,1}
\definecolor{LightOrange}{rgb}{1,0.84,0.80}
\definecolor{LightRed}{rgb}{1,0.50,0.50}
\definecolor{LightGrey}{rgb}{0.9,0.9,0.9}
\definecolor{Green}{rgb}{0.55,0.70,0.0}
\definecolor{LightGreen}{rgb}{0.72,0.91,0.80}
\definecolor{LightYellow}{rgb}{0.94,0.98,0.85}
\newcommand{\ord}[1]{\cellcolor{LightGrey}{#1}}
\newcommand{\best}[1]{\cellcolor{LightGreen}{#1}}
\newcommand{\secBest}[1]{\cellcolor{LightBlue}{#1}}
\newcommand{\better}[1]{\cellcolor{LightBlue}{#1}}
\newcommand{\xhdr}[1]{\vspace{1mm}\noindent{{\bf #1.}}}

\def\SOS{\texttt{SOS}}
\def\EOS{\texttt{EOS}}
\def\rqx{r_{q,x}}
\def\rqxi{r_{q, x_i}}
\def\rqi{r_{q, i}}
\def\hqx{h^{(q,x)}}
\def\hqi{h^{(q,i)}}
\def\hqone{h^{(q,1)}}
\def\hqxi{h^{(q,x_i)}}
\def\rqy{r_{q,y}}
\def\hqy{h^{(q,y)}}
\def\hqyi{h^{(q,y_i)}}
\def\rpy{r_{p,y}}
\def\rpi{r_{p,i}}
\def\rpyi{r_{p,y_i}}
\def\hpy{h^{(p,y)}}
\def\hpyi{h^{(p,y_i)}}
\def\hpyii{h^{(p,y_{i-1})}}
\def\hpii{h^{(p,i+1)}}
\def\rpx{r_{p,x}}
\def\rqyi{r_{q,y_i}}
\def\hpx{h^{(p,x)}}
\def\hpi{h^{(p,i)}}
\def\fqx{f_{q,x}}
\def\fqi{f_{q,i}}
\def\fqone{f_{q,1}}
\def\fqy{f_{q,y}}
\def\fqyi{f_{q,y_i}}
\def\fpx{f_{p,x}}
\def\Mpx{M_{p,x}}
\def\Mpy{M_{p,y}}
\def\Mpi{M_{p,i}}
\def\Mpyi{M_{p,y_i}}
\newcommand{\our}{\textbf{DE-VAE}}
\newcommand{\ourupdate}{gatedUpdate}
\newcommand{\modcyc}{ModCyc}

\title{Fine-grained Sentiment Controlled Text Generation}
\date{}
\author[]{Bidisha Samanta, Mohit Agarwal, Niloy Ganguly}
\affil[]{bidisha@iitkgp.ac.in, mohit2agrawal@gmail.com, niloy@cse.iitkgp.ac.in\\ IIT Kharagpur}

\begin{document}
\maketitle
\begin{abstract} 
Controlled text generation techniques aim to regulate specific attributes (e.g. sentiment) while preserving the attribute independent content. 
The state-of-the-art approaches model the specified attribute as a structured or discrete representation while making the content representation independent of it to achieve a better control.
However, disentangling the text representation into separate latent spaces overlooks complex dependencies  
between content and attribute, leading to generation of poorly constructed and not so meaningful sentences.
Moreover, such an approach fails to provide a finer control on the degree of attribute change. 
To address these problems of controlled text generation, in this paper, we propose \our{}, a hierarchical framework which captures both information enriched entangled representation and attribute specific disentangled representation in different hierarchies. \our{} achieves better control of sentiment as an attribute while preserving the content by learning a suitable lossless transformation network from the disentangled 
sentiment space to the desired entangled representation. 
Through feature supervision on a single dimension of the disentangled representation, \our{} maps the variation of sentiment to a continuous space which helps in smoothly regulating sentiment from positive to negative and vice versa.
Detailed experiments on three publicly available review datasets show the superiority of \our{} over recent state-of-the-art approaches.

\end{abstract}

\section{Introduction}
Text generation using variational inference~\cite{bowman2015generating} is beneficial as it captures important characteristics of the input word sequence in a continuous latent space. The obtained latent space text embedding is broadly used for performing several downstream tasks including machine translation~\cite{bahdanau2014neural}, summarization ~\cite{li2017deep}, dialog generation~\cite{kannan2017adversarial}, etc. In contrast to unconditional text generation, controlled text generative models aim to construct sentences with specified attributes such as sentiment, tense, or style.
Existing state-of-the-art methods for controlled text generation have mainly focused on disentangling the attribute and content representation in the latent space and generate sentences from that by modifying the attribute representation.
Such methods have either derived attribute and content representations separately using multiple attribute specific decoders~\cite{fu2018style,john2018disentangled}, or used an adversarial setup~\cite{hu2017toward} to make the text representation independent of attribute information. Another line of work has focused only on style transfer which primarily involves flipping the style or attribute value of the given sentence.
Several recent works~\cite{logeswaran2018content,shen2017style,singh2018sentiment,zhang2018shaped, zhang2018style,zhao2017adversarially} have explored the use of adversarial discriminators for achieving such text style transfer through disentanglement. However, these methods do not explicitly model the process of fine tuning the
attribute values, e.g., changing sentiment values from extreme positive to neutral to negative, while keeping the content same.
Extension of these models for finer attribute control often fail in mainly two different aspects. They cannot regulate the attribute smoothly because of multidimensional or structured (one hot encoding) representation space of the attributes.
The disentanglement of  the  latent space also overlooks the interdependency between attribute and content. As a result,
just modifying the attribute part  often leads to the generation of sentences
not having sufficient content overlap with the original sentence.

Disentangling latent features (such as rotation and color) of images~\cite{chen2016infogan, higgins2017beta} is a well explored area in computer vision.
In particular, BetaVAE~\cite{higgins2017beta} has modified the Variational Autoencoder with a special emphasis on KL in order to enforce disentanglement.
Further, some other works like~\cite{kim2018disentangling,kumar2017variational} have tried to achieve a trade-off between disentanglement and reconstruction quality.
Although these unsupervised approaches have shown promising results on image data, achieving disentanglement in the latent representation learned for text generation comes with obvious pitfalls.
The difficulty of this task arises from the fact that the text representation based on recent state-of-the-art methods like BERT~\cite{devlin2018bert}, Transformer~\cite{vaswani2017attention}, or Seq2seq~\cite{sutskever2014sequence} is a complex manifold of entangled salient features. These representations are highly expressive for several downstream tasks including high quality sentence generation that preserves the semantics.
Hence, disentangling the enriched representation space into separate spaces for attribute and content results in
neglecting the complex dependency between them,
thereby compromising the quality of text generation.
Thus, previous methods ~\cite{hu2017toward,john2018disentangled,logeswaran2018content,shen2017style,singh2018sentiment,zhang2018shaped,zhang2018style,zhao2017adversarially} that have tried to achieve attribute control by disentangling the latent space used for text generation suffer from these aforementioned issues, i.e., generating poor and unrealistic sentences while trying to fine tune the attribute values.
A recent work~\cite{wang2019controllable} has proposed attribute transfer using entangled representation that enables fine tuning of the sentiment polarity. However, it is restricted to modifying the sentiment of the given sentence to only its opposite polarity using costly Fast-Gradient-Iterative Modification.
We aim to address these shortcomings of existing literature by proposing a trade-off between preserving entangled representation as a latent space to generate text for better reconstruction, and deriving a continuous attribute representation in a disentangled space on top of it
to fine tune the attribute values.





In this paper, specifically, we propose the model \emph{Disentangled-Entangle-VAE}~(\our{}), a feature supervised
framework that transforms  entangled and enriched text representation obtained using \emph{BERT} encoder to a higher level representation of sentiment as a specified attribute, along with other unspecified
attributes using a transformation network.
\our{} enforces disentanglement of the derived representation
by imposing a factored prior
which enables 
independence among different dimensions of the latent representation.
Further, using attribute supervision on the intended dimension of this disentangled representation, we map the sentiment to a continuous space.
For attribute guided text generation, \our{}  converts the disentangled representation back to original entangled representation and generate sentence from that.
This reverse transformation  helps to preserve the complex relationships between sentiment and other inherent attributes in an enriched entangled space, thereby generating more meaningful and realistic sentences compared to competing methods.
However, the choice of such transformation networks for transforming from disentangled to entangled space, and vice-versa is extremely important.
The transformation needs to be lossless,
otherwise the decoded  entangled space can become different than that of the
original text representation.
We use the concept of invertible normalising flow~\cite{dinh2016density,kingma2016improved,rezende2015variational} to  enforce these transformations.
By jointly optimizing the parameters of  transformation network, along with
imposing disentanglement constraint and feature supervision on the feature space, \our{} successfully learns how a disentangled 
space for sentiment and other unspecified attributes can be converted to a meaningful entangled representation space of the sentence.





We demonstrate the effectiveness of \our{} to generate controlled text by fine tuning sentiment.
Using three large publicly available review datasets, we show that \our{} improves the performance significantly
over previous controlled text generative models on three different criteria, namely, sentiment control accuracy,  
smooth fine tuning of sentiment in both directions,
 and content preservation.
We show that even a small difference between the decoded entangled feature space and the original entangled space, introduced by using an alternative transformation network, can lead to a significant performance drop. Finally, through an ablation study, we demonstrate the disentanglement
achieved by \our{} on the derived feature space.

\begin{figure*}
	\centering
	{\includegraphics[clip,width=0.85\columnwidth,height=40mm]{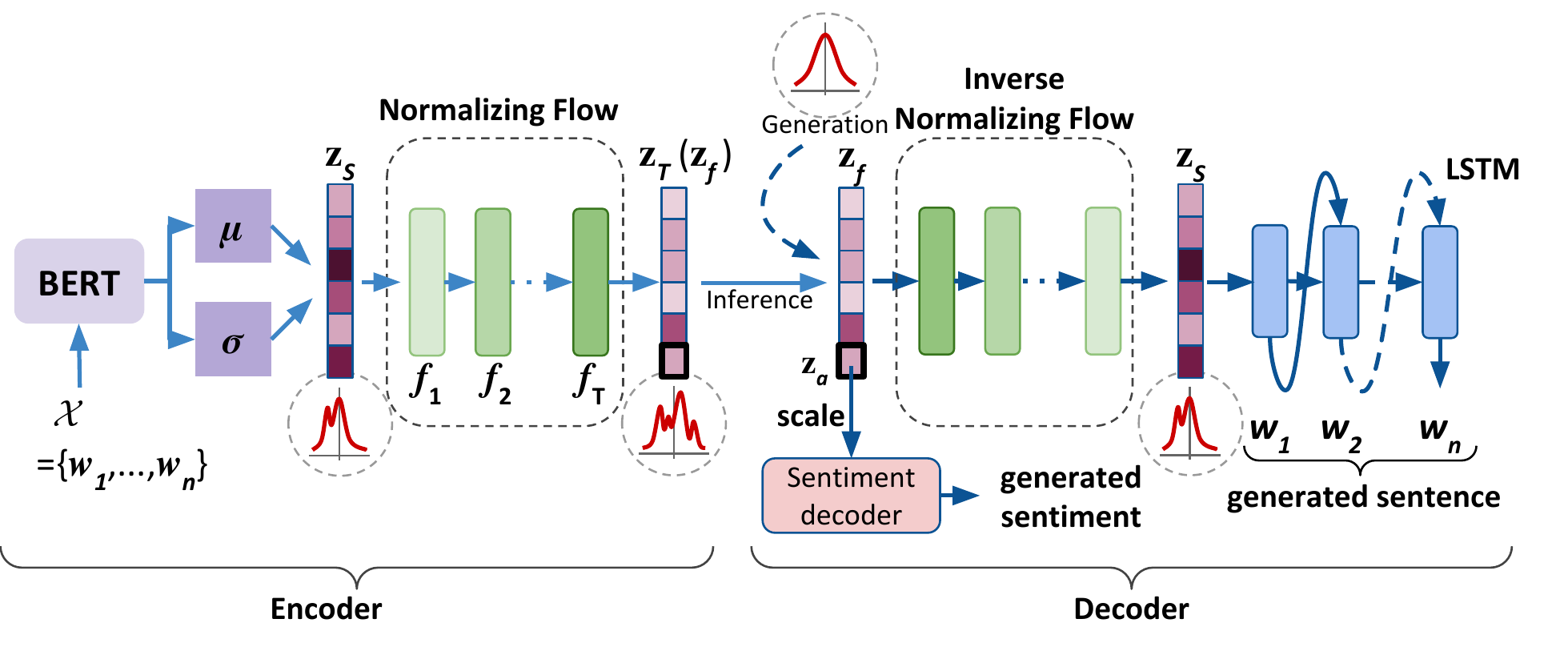}}
	\caption{The Encoder and Decoder module of \our{}. The encoding process takes a word sequence $\xb$ and obtain its BERT embedding. Then it converts it to a continuous representation $\zb_s$ followed by
	transformation to a feature representation $\zb_f$ using invertible normalising flow. It assigns the last dimension of $\zb_f$ for sentiment representation $\zb_a$. The decoding process samples $\zb_f$ from prior or posterior. It decodes sentiment from the  $\zb_a$. 
	Then the feature representation $\zb_f$ is transformed to the sentence representation $\zb_s$ via an inverse flow. $\zb_s$ is then used to generate the word sequence $\xb$.
	}
	\label{fig:architecture}
\end{figure*}

\section{\our{} : Text generation with fine tuned attributes}
In this section, we first give a high level overview of \our{}, a hierarchical model
for controlled text generation, starting with describing the data.
Next, we describe the key
technical aspects of its individual components with key contributions in detail. Finally, we illustrate the training procedure.
%
\subsection{Model overview}
We consider an input set $X = \{\xb_0,\cdots, \xb_{M-1}\}$ of $M$ observed sentences sampled from some underlying unknown data distribution $p_D$.
Along with the sentences, we have the corresponding ground truth observed attribute, sentiment,
denoted as
$F = \{f_0,\cdots,  f_{M-1}\}$. Here $f_i$ is associated to sentence $\xb_i$. For ease of reference, we will henceforth denote a training instance $\xb_i$ and $f_i$ by $\xb$ and $f$ respectively.
Detailed architectural overview of \our{} is shown in Figure ~\ref{fig:architecture}. The whole architecture can be divided into two modules consisting of a hierarchical encoder and a corresponding hierarchical decoder.
We start by describing the inference model (encoder) followed by the  generation model (decoder).
\subsection{Inference model}
The inference model is designed as a bottom-up hierarchical encoder. 
It has two distinct layers for modeling word sequence representation $\zb_s$, and derived feature representation $\zb_f$.
The posterior distribution of our hierarchical inference mechanism can be represented as a factored model like:
\begin{align}
\label{eq:encoder}
q_{\phi}(\zb | \xb ) =
&\underbrace{q_{\phi}(\zb_{s} | \xb)}_{\text{Entangled}}\underbrace{q_{\phi}(\zb_{f} | \zb_s)}_{\text{Disentangled}}
\end{align}
We design an entangled sentence encoder $q_{\phi}(\zb_{s} | \xb)$ in the lowest layer as follows.
Given a word sequence $\xb$, we obtain the word embeddings $\Ecal_w$ for each word $w$ in $\xb$ from the BERT pre-trained model~\cite{turc2019}.  Then we combine these representations to generate an aggregated sentence encoding $\Ecal_s$ and transform it into a continuous Gaussian space $\zb_s \in \RR^d$, as follows:
\begin{align}
&\Ecal_w = \{\eb_1,\dots, \eb_{|\xb|}\} = \text{BERT} (\xb),  \Ecal_s =\frac{1}{|\xb|} \sum_{\eb \in \Ecal_w} \eb \notag \\
&q_{\phi}(\zb_s | \xb) =  \Ncal(\mu_s, \diag(\sigma_s^2))\ \text{where}\ [\mu_s, \sigma_s] = g_\phi (\Ecal_s)
\end{align}
We transform
$\Ecal_s$ to a Gaussian distribution parameterized by $[\mu_s, \sigma_s]$ using a fully connected neural network $g_\phi$. The sentence representation $\zb_s \in \RR^d$ is an entangled representation with $d$ dimensions. Next, we aim to  transform this representation $\zb_s$ to another representation $\zb_f \in \RR^d$ on which we can impose feature supervision and disentanglement for  attribute control.  Design of $q_\phi(\zb_f|\zb_s)$ is crucial for fine-tuned attribute control.  First, it should transfer all the necessary information of $\zb_s$ to $\zb_f$ so that 
we can leverage the information for sentiment supervision
on $\zb_f$. Most importantly, the transformation needs to be invertible. As we are interested to control the attribute by modulating  $\zb_f$, and then transforming back to
 $\zb_s$ distribution space, any transformation other than an invertible one will lead to some different distribution space causing important information loss.
Hence, we use an invertible normalizing flow~\cite{dinh2016density,kingma2016improved,rezende2015variational} to design $q_\phi(\zb_f | \zb_s)$.

A normalizing flow is a powerful transformation function which applies a chain of invertible parametrized transformations $\textbf{f}_t (t = 1,\dots,T)$ to its input (here $\zb_s$) such that the outcome of the last iteration, $\zb_T$, has a more flexible distribution (here $\zb_f$). We have used an effective autoregressive transformation flow R-NVP~\cite{dinh2016density},
which copies the first $k$ dimensions ($1<k<d$) of the input, while shifting and scaling all the remaining ones. Specifically,
the estimated approximate transformation flow $\textbf{f}_t$, i.e. $q_t(\zb_t | \zb_{t-1})$, can be characterized as:
\begin{align}
{\zb_t}_{(1:k)} = {\zb_{t-1}}_{(1:k)}, \ \text{and}
\ {\zb_t}_{(k+1:d)} =  {\zb_{t-1}}_{(k+1:d)} \cdot \sigma_t + \mu_t, \text{where}\ [\mu_t, \sigma_t]  = \Psi_t({\zb_{t-1}}_{(1:k)})
\label{eq:flow}
\end{align}
Here, $\Psi_t$ are designed as multilayer fully connected feed-forward networks which are not invertible 
 However, a careful inspection of Eq~\ref{eq:flow} reveals that given $\zb_t$, the input $\zb_{t-1}$ can be fully recovered. So this makes the transformation flow $\textbf{f}_t$ invertible. Thus, we can write $q_{\phi}(\zb_{f} | \zb_s) := q_{\phi}(\zb_T | \zb_s)$ and we assign $\zb_f := \zb_T$.
From the characterization of R-NVP, we observe that ${\zb_f}_{(k+1:d)}$ will have the aggregated information coming from all dimensions of $\zb_s$. So we pick the $d^{th}$ (last) dimension of $\zb_f$ as  $\zb_a$ to further fine-tune it using sentiment supervision which we will discuss in detail in the next section. The rest of the dimensions of $\zb_f$ are kept for unspecified features denoted by $\zb_u$.
We will discuss how we achieve the disentanglement of $\zb_f$ in Sec.~\ref{sec:train} while discussing the training objective.

\subsection{Generative model}
\label{sec:model}
 We design our generative model $p_\theta$ using a top-down hierarchy, with
 two different variables $\zb_s$ and $\zb_f$. The overall distribution of the latent variables is defined as:
\begin{align}
\label{eq:latents}
p_\theta(\zb) =  & \underbrace{p_\pi(\zb_f)}_{\text{Disentangled}} \underbrace{p_\theta(\zb_s | \zb_f)}_{\text{Entangled}}
\end{align}
Here $p_\pi(\zb_f)$ is a factored prior of the feature representation $\zb_f$, which can be expressed as
$\zb_f
= \prod_{i =1}^{d} p_\pi(\zb_f^i)$.
As discussed in the previous section, we have designated the last dimension of the disentangled attribute representation to capture sentiment,  
and remaining dimensions for unspecified features. Henceforth,  sentiment representation can be sampled from $\zb_a \sim p_\pi(\zb_f^d)$ and unspecified representations can be sampled as $\zb_u \sim \prod_{i =1}^{d-1} p_\pi(\zb_f^i)$.
The  factorized prior distribution $p_\pi(\zb_f)$ is designed to be standard normal distribution, hence $p_\pi(z^i_f) \sim \Ncal(0, I) \ \forall i $ in $[1,d]$.
To facilitate smooth interpolation in sentiment space, which is one of the major differences of \our{} than other alternatives, we use feature supervision on $\zb_a$ as follows. Given $\zb_a$, we try to decode the sentiment of the given sentence $\xb$ and back propagate the classification error to modify the values of $\zb_a$.
More specifically, the decoding distribution for the ground truth sentiment is represented as:
\begin{align}
\label{eq:featuresample}
p_\theta(\fb | \zb_a) = \text{Categorical}(\xi (\zb_a))
\end{align}
Here $\xi$ is a scaling network to convert the single value $\zb_a$ into a two dimensional logit to calculate the likelihood of ground-truth sentiment.
Next, the network tries to decode the entangled distribution $\zb_s$ from the disentangled distribution $\zb_f$. As described in Eq.~\ref{eq:flow}, we apply the reverse transformation flow to recover $\zb_s$ using T inverse flows $\textbf{f}_t^{-1} (t = 1,\dots,T)$. 
Starting from $\zb_T$, denoted by $\zb_f$, we obtain $\textbf{f}_t^{-1}$, i.e. $p_t(\zb_{t-1} | \zb_{t})$, described using the inverse transformation flow below:
\begin{align}
{\zb_{t}}_{(1:k)}= {\zb_{t+1}}_{(1:k)}, \ \text{and}
\ {\zb_t}_{(k+1:d)} = \frac{{\zb_{t+1}}_{(k+1:d)} - \mu_t}{ \sigma_t} , \text{where}\ [\mu_t, \sigma_t]  = \Psi_t({\zb_{t}}_{(1:k)})
\label{eq:flow_rev}
\end{align}
It may be noted that $\mu_t$ and $\sigma_t$ are derived from the neural network $\Psi_t$ which is shared between the encoder and decoder. The log probability density of $p_\theta(\zb_s | \zb_f)$, i.e. $\text{log}\ p_T(\zb_s|\zb_f),$~\cite{dinh2016density} becomes equivalent to:
\begin{align}
\text{log}\ p_T(\zb_s|\zb_f) = \text{log}\ p_\pi(\zb_f) - \sum_{t=1}^{T} \text{log det}\frac{d\textbf{f}_t}{d \textbf{f}_{t-1}}
\label{eq:flowlikelihood}
\end{align}

Finally, with the decoded $\zb_s$, we sample the word sequence as follows:
\begin{align}
\label{eq:wordsample}
&h(j) = r_\theta(x(j-1),  \zb_s) \text{ and }
x(j) \sim \text{Softmax}(m_\theta(h(j)))
\end{align}
Here $r_\theta$ is a gated recurrent unit, which takes the previously generated token $x(j-1)$ and the sentence representation $z_s$ to generate the hidden state $h(j)$. Then we pass this hidden state information to a feedforward network $m_\theta$ to generate logits. Subsequently, we sample words based on the softmax distribution of the generated logits.
 We define the joint likelihood of the sentence, features and the latent variables as:
\begin{align}
\label{eq:likelihood}
p_\theta(\xb, \fb, \zb_s,  \zb_f) &
= p_\theta(\xb|\zb_s) p_\theta(\fb| \zb_f)p_\theta(\zb_s| \zb_f) p_\pi(\zb_f)
= p_\theta(\xb|\zb_s) p_\theta(\fb| \zb_a)p_\theta(\zb_s| \zb_f) p_\pi(\zb_f)
\end{align}

\subsection{Training}
\label{sec:train}
Instead of optimizing the joint likelihood given in Eq.~\ref{eq:likelihood} by training both layers of the hierarchy simultaneously, we train \our{} in two phases. First, we train the lower layer which is responsible for sentence reconstruction; next, we train the transformation flow network as well as the upper layer.
We train the lower layer by maximizing the marginal likelihood of the input sentence $\xb$ as follows:
\begin{align}
\text{log}\ p_\theta(\xb)
&\ge \EE_{q_\phi(\zb_s| \xb)} \text{log}\ p_\theta(\xb|\zb_s)  - \text{KL}(q_{\phi}(\zb_s|\xb) ||  p_\pi(\zb_s))
\end{align}
Next, we update the flow parameters (Eq. \eqref{eq:flowlikelihood}) and impose feature supervision
by maximizing the lower bound of marginal likelihood of $\zb_s$ which is $\text{log}\ p_\theta(\zb_s)$ :
\begin{align}
 \EE_{q_\phi(\zb_f| \zb_s)} \Big [  \text{log}\  p_\theta(\fb|\zb_a) + \text{log}\ p_\pi(\zb_f) - \sum_{t=1}^{T} \text{log det}\frac{d\textbf{f}_t}{d \textbf{f}_{t-1}} \Big ]
- \text{KL}(q_{\phi}(\zb_f | \zb_s)|| p_\pi(\zb_f))
\end{align}

We may further breakdown the KL term of the above objective function by taking an expectation over $\zb_s$
as following:
\begin{align}
\EE_{\zb\sim q_\phi(\zb_s)} I(\zb_s, \zb_f) +\underbrace{ \text{KL} (q_{\phi}(\zb_f | \zb_s) || p_\theta(\zb_f))}_{\text{Total Correlation}}
\end{align}
As $p_\pi(\zb_f)$ is fully factorized, minimizing the above total correlation loss will benefit the model in achieving disentanglement of $\zb_f$ along the dimensions\cite{higgins2017beta}. 
It is also important that  a specified disentangled part of $\zb_f$, for a designated sentiment $f$, should carry enough information about that sentiment. Hence, the mutual information between the sentiment $f$ and  $\zb_a$  should be high. This mutual information
can be computed using entropy function $H(.)$ as:
\begin{align}
\label{eq:MI}
I(f, \zb_a) = H(f) - H(f | \zb_a) &
  \ge \EE _{\xb\sim p_D} [\EE_{q_\phi(\zb_s | \xb) q_\phi(\zb_a | \zb_s)}\text{log} \ p_\theta(f | \zb_a ) ]
\end{align}
As $I(f, \zb_a)$ is lower bounded by the likelihood $p_\theta(f | \zb_a)
$, we provide extra emphasis of the likelihood term in the objective function.
To optimize the upper layer and flow parameters we maximize the lower bound of $\text{log}\ p_\theta(\zb_s)$ as below:
\begin{align}
\label{eq:up}
\EE_{q_\phi(\zb_f| \zb_s)} \Big [\beta \text{log}\  p_\theta(\fb|\zb_a) + \text{log}\ p_\pi(\zb_f) - \sum_{t=1}^{T} \text{log det} \frac{d\textbf{f}_t}{d \textbf{f}_{t-1}} \Big ] - \gamma \text{KL}(q_{\phi}(\zb_f | \zb_s)|| p_\pi(\zb_f))
\end{align}
where $\beta$ and $\gamma$ are regularizing parameters to emphasize on increasing the sentiment likelihood and enforce the disentanglement of $\zb_f$. Updating the flow parameters along with high emphasis on disentanglement and feature supervision using Eq ~\ref{eq:up} helps the model to learn the complex transformation of independent features $\zb_u$ and $\zb_a$ to an enriched entangled $\zb_s$. The specific details of implementation is provided in the Appendix A.
\section{Experimental evaluation}
In this section, we evaluate the performance of \our{} in terms of sentiment control using three different evaluation criteria -
(a) sentiment control accuracy,
(b) fine tuning of sentiment, and (c) content preservation.
Additionally, we also discuss the
extent of disentanglement in the latent space $\zb_f$.

 For our evaluation, 
 we rely on three large review datasets for sentiment controlled text generation - (a) \textbf{Yelp}~\cite{wang2019controllable} with $443248$, $2000$, and $1000$ number of labeled data on restaurant review for training, validation, and test respectively with a vocabulary size of $9.5$K, (b) \textbf{Amazon}~\cite{wang2019controllable} with $554997$, $2000$, and $2000$ number of labeled data on product review for training, validation, and test respectively with a vocabulary size $25$K, and (c) \textbf{IMDB}~\cite{diao2014jointly} movie review corpus which consists of $708929$, $4000$, and $2000$ number of unlabeled training, validation, and test data with a vocabulary size of $28$K. This IMDB data has been tagged by Stanford sentiment tagger~\cite{socher2013recursive} and converted to three sentiment labels, namely, positive, neutral, and negative. 

We compare the performance of \our{} with the following baselines that focuses on controlled text generation using disentanglement - 
(a) \textbf{ctrlGen}~\cite{hu2017toward} which is a semi-supervised method for sentiment oriented text generation, and (b)~\textbf{DAE}~\cite{john2018disentangled} which is a supervised method that focuses on disentanglement using adversarial loss.
Other than these, we also compare \our{} with \textbf{entangleGen}~\cite{wang2019controllable} which focuses on text style transfer using entangled representation. Apart from these state-of-the-art baselines, we also use \textbf{DE-VAE-NR} (\our{} \textbf{N}on-\textbf{R}eversible transformation) as a baseline which is a variation of \our{}. In \textbf{DE-VAE-NR}, we replace the invertible normalizing flow used in \our{} with two neural networks designed as a two layer fully connected feedforward network responsible for capturing $q_\phi(\zb_f | \zb_s)$ and $p_\theta(\zb_s | \zb_f)$. 
\begin{table}[]
	\small
	\centering
		\scalebox{0.8}{
	\begin{tabular}{c|c|c|c|c|c|c}
		\hline
		& \multicolumn{2}{c|}{Yelp} & \multicolumn{2}{l|}{Amazon} & \multicolumn{2}{l}{IMDB} \\ \hline
		
		Methods &\multicolumn{1}{l|}{\begin{tabular}[c]{@{}l@{}} Controlled\\generation \end{tabular}} &
		\begin{tabular}[c]{@{}l@{}}Text style\\  transfer\end{tabular} &
		\multicolumn{1}{l|}{\begin{tabular}[c]{@{}l@{}} Controlled\\generation  \end{tabular}} &
		\begin{tabular}[c]{@{}l@{}}Text style\\  transfer\end{tabular} &
		\multicolumn{1}{l|}{\begin{tabular}[c]{@{}l@{}} Controlled\\generation \end{tabular}} &
		\begin{tabular}[c]{@{}l@{}}Text style\\  transfer\end{tabular} \\ \hline
		
		ctrlGen     &      0.72  &    0.70&         0.62             &0.63    &0.76                 &0.77   \\ \hline
		DAE         &    0.95        &  0.80   &   0.84   &   0.77 &    0.82         &  0.81 \\ \hline
		entangleGen &     -     &   \textbf{0.94} &     -        &  0.82   &    -         & 0.66  \\ \hline
		
		\hline
		\textbf{DE-VAE-NR} & 0.62       & 0.58   &   0.59       &0.51    & 0.59   &0.53   \\ \hline
		\our{}         &  \textbf{0.95}       & \textit{0.84}    &   \textbf{0.84}           &  \textbf{0.90}  &   \textbf{0.90}           & \textbf{0.86}  \\ \hline
	\end{tabular} }
	\caption{Controlled generation and Text style transfer accuracy achieved by different methods. }
	\label{tab:sentiment}
\end{table}
\subsection{Sentiment control accuracy}
Here we measure the quality of sentiment control by quantitatively evaluating the sentiment oriented sentence generation accuracy.
For this purpose, we train a sentiment classifier by extending \textbf{BERT}~\cite{devlin2018bert}. This classifier is $95\%$ accurate on Yelp data and $85\%$ accurate on Amazon and IMDB data
 which demonstrates its robustness. 
Specifically, \our{} generates sentences by regulating the values in the designated dimension of $\zb_f$ which is modeled to control the sentiment; we use the pre-trained sentiment classifier to assign sentiment labels to the generated sentences. Similarly, we generate sentences using the baseline methods.
We report the accuracy of sentiment oriented sentence generation 
in two different ways - (a) \textbf{Controlled generation accuracy} which is accuracy of generating sentences where the generative representation is sampled from the prior (i.e., from $p_\pi(\zb_f)$ in case of \our{}), followed by assigning the desired value for the sentiment representation (i.e., $\zb_a$ for \our{}), and
(b)  \textbf{Text style transfer accuracy} which is accuracy of generating sentences with opposite polarity sentiment from the representation 
of a sentence $\xb$, 
bearing a specific type of sentiment (in our case, $\zb_f \sim q_\phi(\zb_f | \zb_s) q_\phi(\zb_s | \xb)$, as it regulates sentiment).

From the results reported in Table~\ref{tab:sentiment}, it can be observed that \our{} outperforms all competing methods across all datasets for controlled text generation, other than text style transfer accuracy in case of Yelp. 
The superior performance of \our{} stems from the fact that it learns a disentangled feature representation where sentiment information is modeled in a single dimension which is independent of other dimensions. So, regulating the sentiment value along this designated dimension gives better control while generating sentences bearing the same sentiment. 
Further, in case of \textbf{DE-VAE-NR}, we train the parameters of the transformation network by maximizing the likelihood of $\zb_s$, i.e. $\EE_{q_\phi(\zb_s | \xb)}p_\theta(\zb_s | \zb_f)$, such that the minimum KL between $q_\phi(\zb_f | \zb_s)$ and $p_\theta(\zb_s | \zb_f)$ is found to be in the range $0.2-0.4$ 
across all datasets; however, the performance of \textbf{DE-VAE-NR} is significantly inferior compared to \our{}.
Close inspection reveals that even though the KL was low, as the decoded distribution $\zb_s$ was not exactly the same as the encoded distribution, it was generating sentences very different from the original sentences. This shows the importance of invertible normalizing flow as a design choice. 

The closest competitor to our method is \textbf{DAE} for controlled text generation with comparable performance to \our{}. 
The reason is that \textbf{DAE} achieves disentanglement between style space and content space using extensive adversarial training and incorporates auxiliary multi-task loss for sentiment or style.
 Further, in comparison to other methods, \textbf{ctrlGen} performs poorly for controlled text generation. 
 We speculate that \textbf{ctrlGen} has comparatively less control than other methods since it does not model feature representation from sentence representation and tries to control generation by providing only a one hot encoding of sentiment externally.
  We have reported the accuracy of \textbf{entangleGen} only for text style transfer in Table~\ref{tab:sentiment} since it focuses only on attribute transfer. We observe that \textbf{entangleGen} achieves better accuracy on Yelp dataset than our method; however, \our{} performs better on all other datasets.
  Since \textbf{entangleGen} inherently supports binary-valued attributes by design,
 it can best modify the sentiment value towards its opposite polarity 
  using Fast-Gradient-Iterative-Modification. 
Hence, its performance suffers in IMDB data where there are more than two sentiment values. 
\begin{figure}[!t]
	\vspace{-4mm}
	\centering
		\hspace*{-0.55 cm}
		\includegraphics[width=0.34\textwidth, clip]{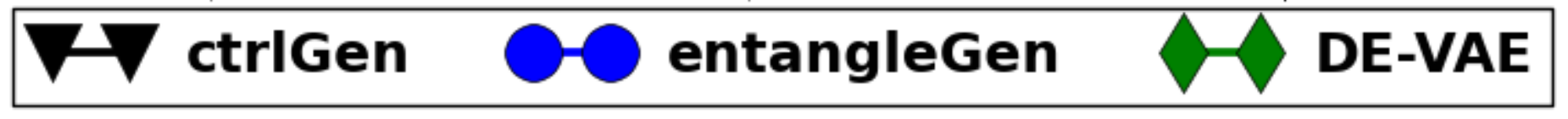} 	\vspace*{-0.30 cm} \\
	\hspace*{-0.55 cm}
	\subfloat[Yelp (Positive) ]{\includegraphics[width=0.25\textwidth, clip]{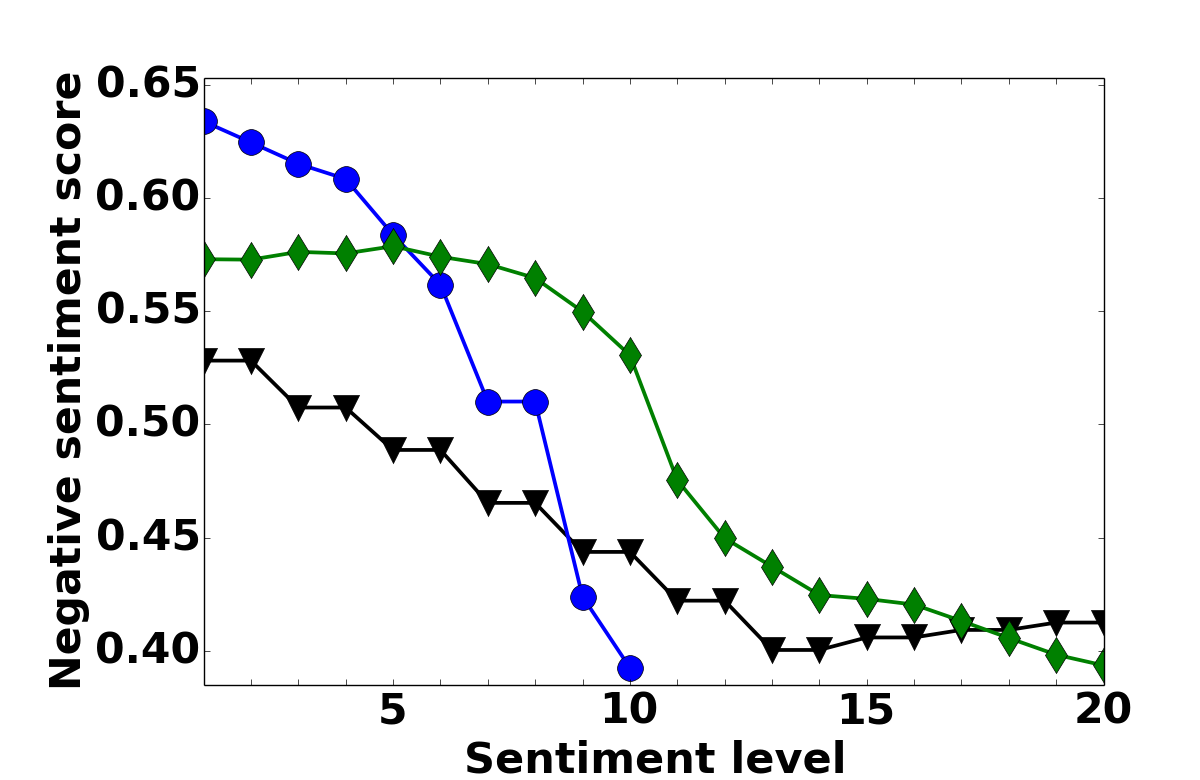}}
	\hspace*{-0.3 cm}
	\subfloat[Amazon (Positive)]{\includegraphics[width=0.25\textwidth, clip]{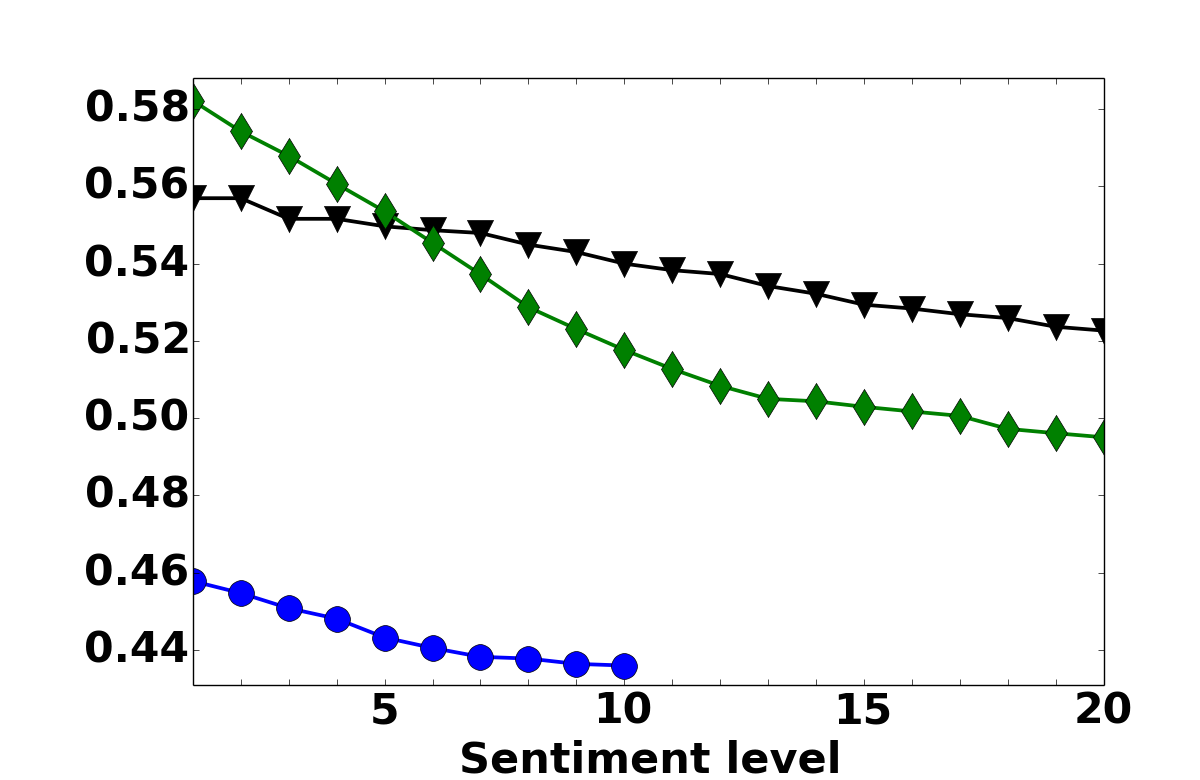}}
	\hspace*{-0.30 cm}
	\subfloat[Yelp (Negative)]{\includegraphics[width=0.25\textwidth, clip]{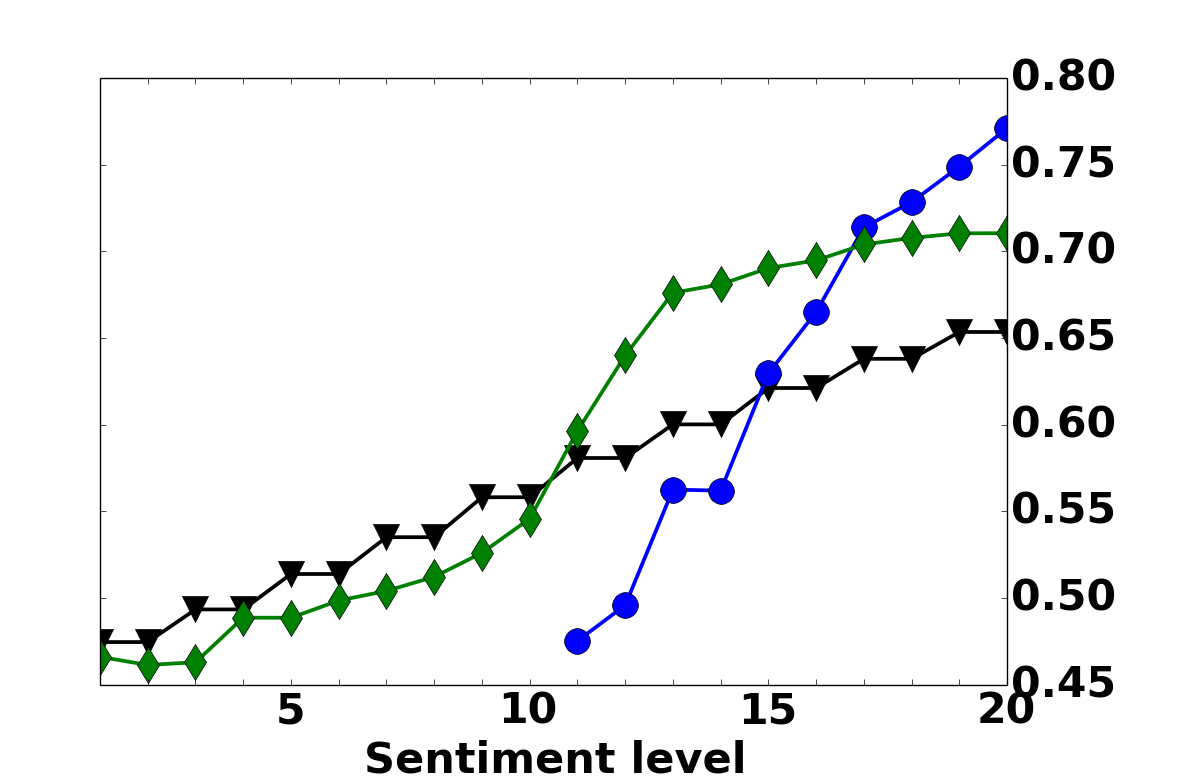}}	
	\hspace*{-0.3 cm}
	\subfloat[Amazon (Negative)]{\includegraphics[width=0.25\textwidth, clip]{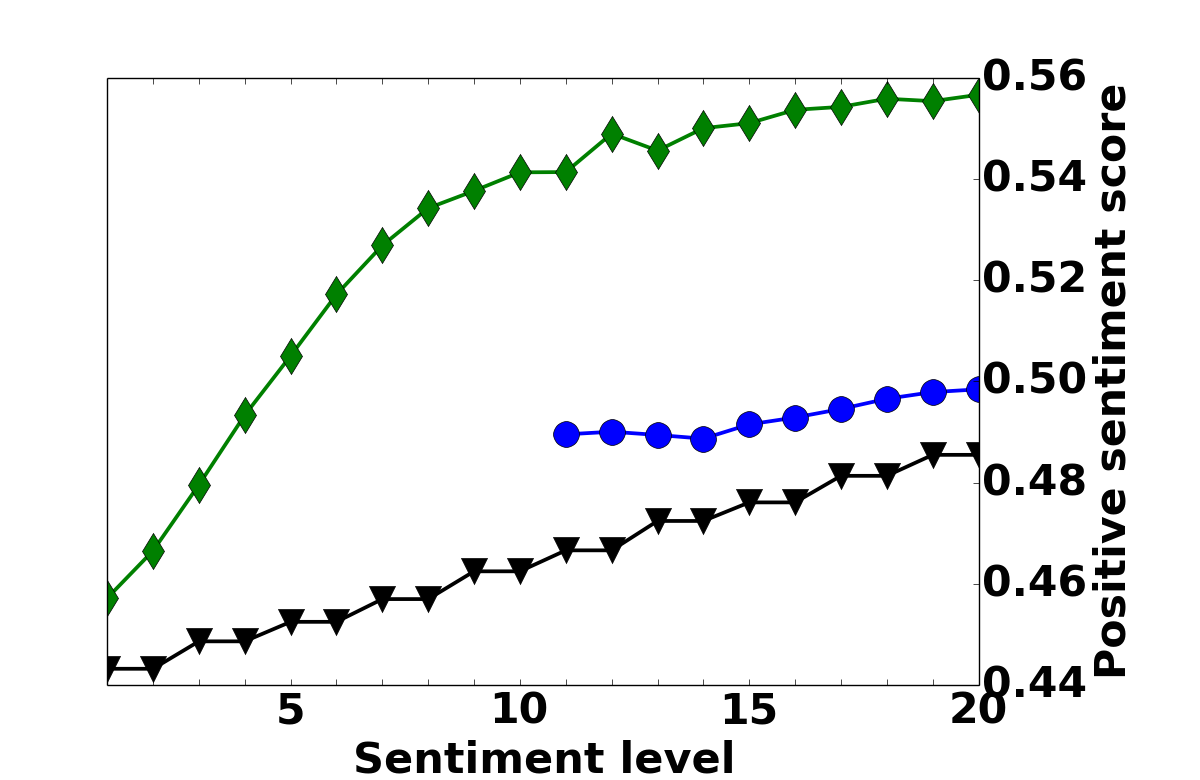}}\\
	\hspace*{-0.55 cm}
	\subfloat[Yelp (Positive)]{\includegraphics[width=0.25\textwidth, clip]{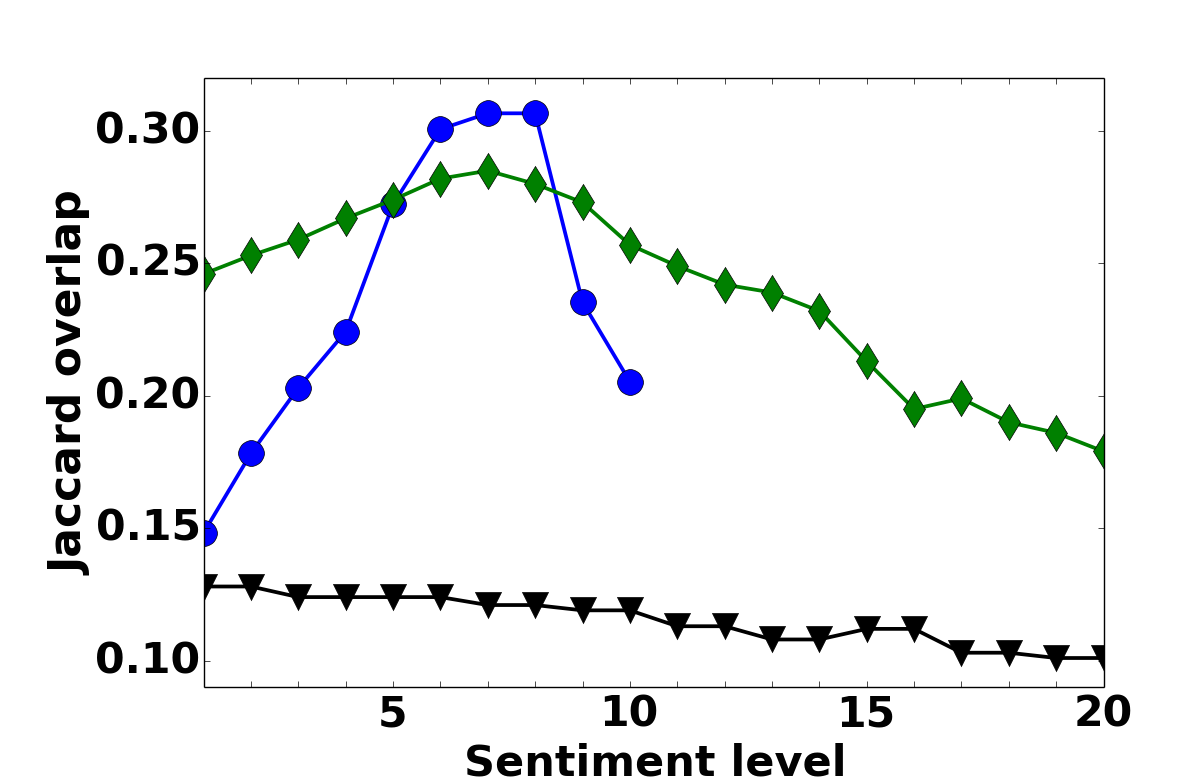}}
	\hspace*{-0.3 cm}
	\subfloat[Amazon (Positive)]{\includegraphics[width=0.25\textwidth, clip]{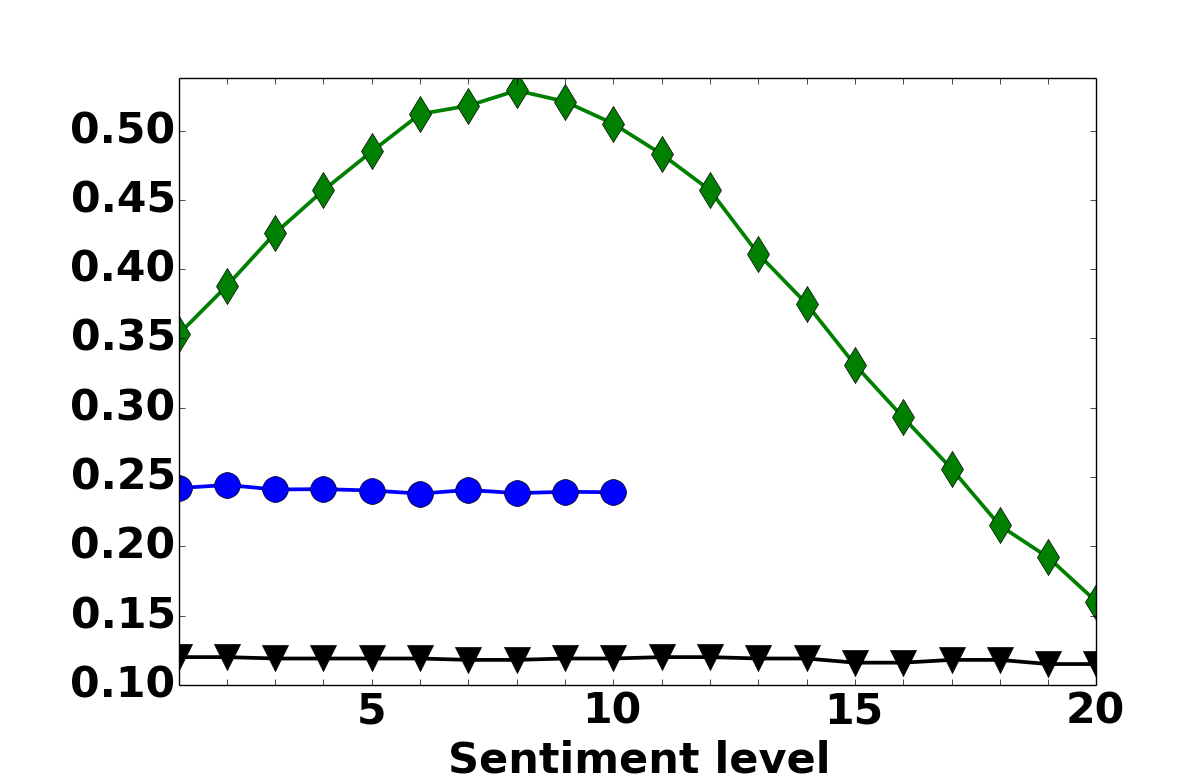}}
	\hspace*{-0.30 cm}
	\subfloat[Yelp (Negative)]{\includegraphics[width=0.25\textwidth, clip]{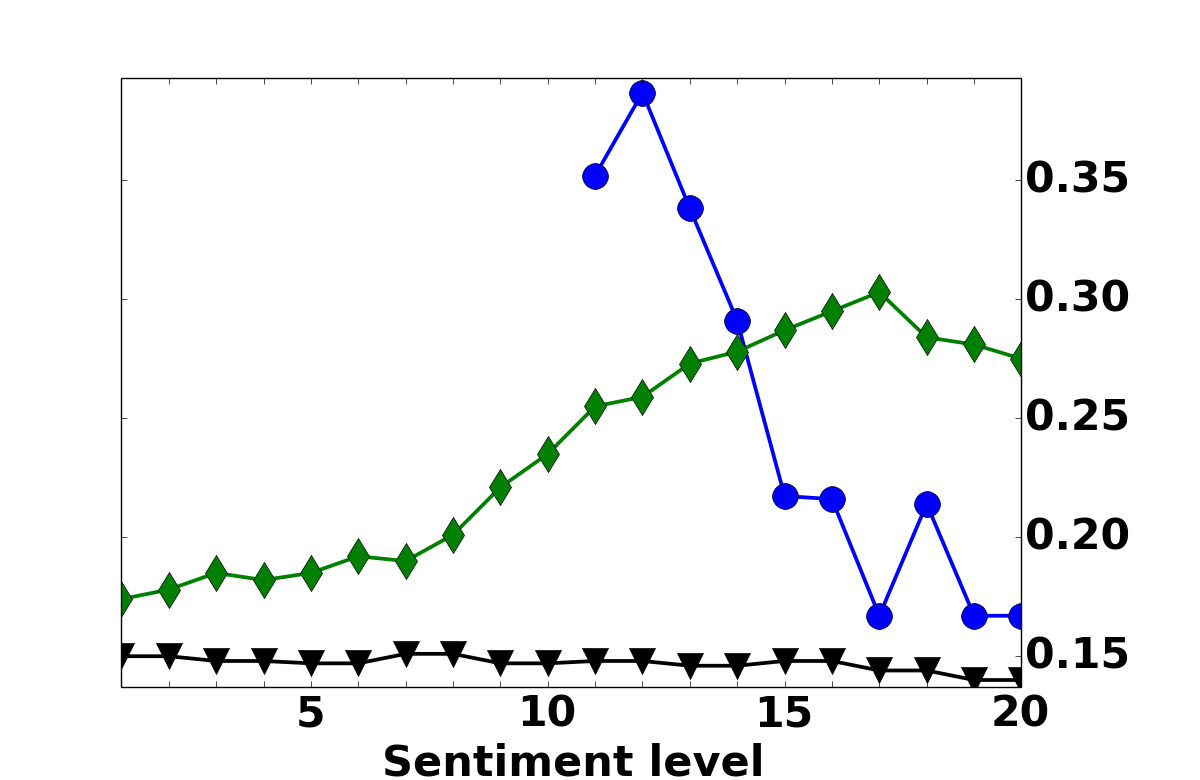}}	
	\hspace*{-0.3 cm}
	\subfloat[Amazon (Negative)]{\includegraphics[width=0.25\textwidth, clip]{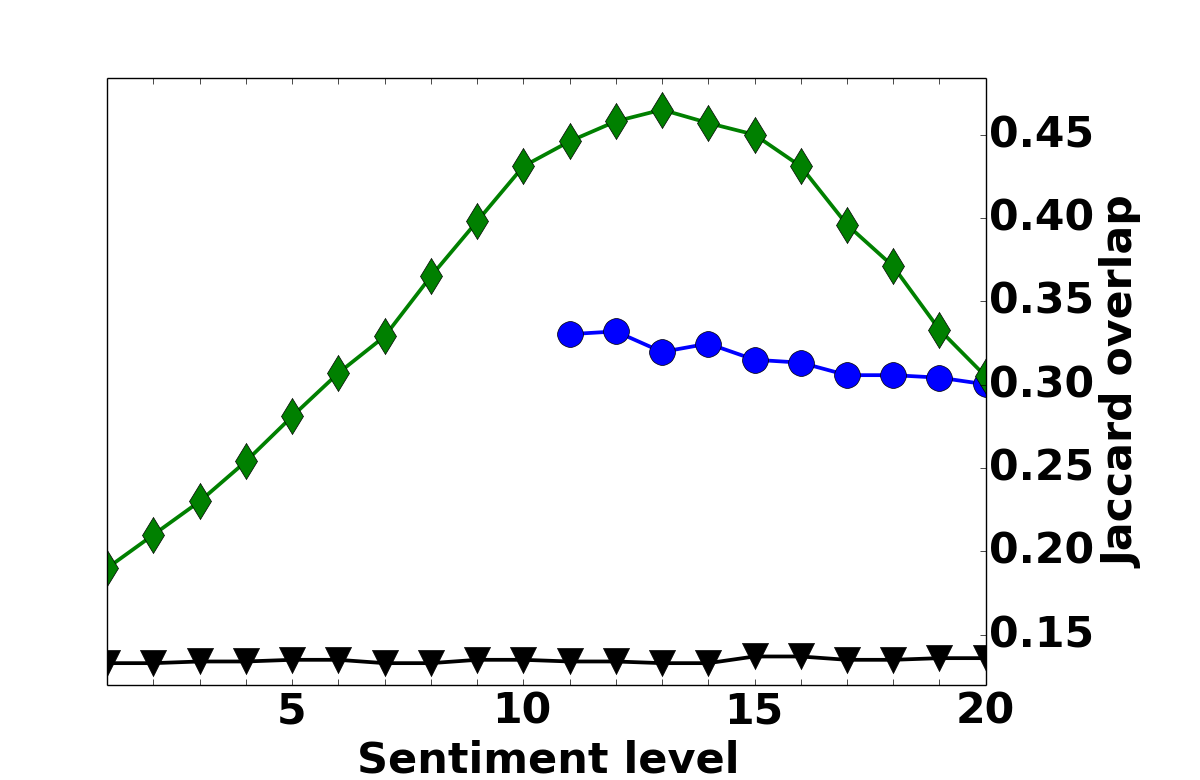}}
	\caption{Upper row denotes variation of average sentiment score of generated sentences with fine-tuned degree of sentiment control. Bottom row indicates variation of  Jaccard score of generated sentences with fine-tuned degree of sentiment control. The original sentence's sentiment is in brackets.
	}
	\vspace*{-0.2cm}
	\label{fig:labelvariation}
\end{figure}
\subsection{Fine tuning of sentiment and Content preservation}
In this section, we discuss the performance of \our{} in terms of both the evaluation criteria - fine tuning of sentiment, and content preservation. 
As by design, \our{} is able to capture the entire behaviour of the sentiment in a continuous latent space of a single dimension, we first estimate the maximum value $f_{max}$ and the minimum value $f_{min}$ of this dimension from the training data.
Then we use this range and
interpolate between $f_{max}$and $f_{min}$ in $20$ different sentiment levels where the value of $i^{th}$ level $l_i$ ($1\leq i\leq 20$) is denoted as $f_{min} + \frac{(f_{max} - f_{min}) (i-1)}{19}$. Given a feature representation $\zb_f$ corresponding to a sentence $\xb$, we assign a level to $\zb_a$ (last dimension of $\zb_f$) and sample $100$ sentences corresponding to this modified $\zb_a$. We repeat this procedure
corresponding to each level. 
Similarly, we extend \textbf{ctrlGen} to generate sentences by interpolating $20$ different values sampled from the range $[0,1]$ for the structured two-dimensional sentiment representation.
As \textbf{entangleGen} can only regulate the sentiment of the given sentence in the opposite polarity
using $35$ different degrees of sentiment, 
we have aggregated them within ten sentiment levels. It may be noted that we have excluded \textbf{DAE} from this evaluation since it models sentiment in a multi-dimensional space and fine tuning sentiment in such a space is non-trivial
and needs separate investigation. Additionally, we have also excluded comparison with \textbf{DE-VAE-NR} due to its poor performance on sentiment control.   

To quantitatively measure the performance of fine tuning of sentiment, we have used the pre-trained Stanford sentiment regressor~\cite{socher2013recursive} which provides sentiment scores of sentences generated for \our{} and other methods between $[0,1]$ with $1$ denoting most positive and $0$ denoting most negative. 
Further, we evaluate the model performance in terms of content preservation while fine tuning sentiment. 
For this purpose, we compute \emph{Jaccard overlap}~\cite{tustison2009introducing} of unigram words in the original sentence  $x$ and the generated sentence $y$, computed as $\frac{|w_x \cap w_y|}{|w_x \cup w_y|}$, where $w_{.}$ denotes the set of words after excluding stopwords in corresponding sentence. 

In Figure~\ref{fig:labelvariation}, we have demonstrated the variation of mean sentiment scores (Figures~\ref{fig:labelvariation}(a)-(d)) and \emph{Jaccard overlap} scores (Figures~\ref{fig:labelvariation}(e)-(h)) for different $l_i$ (with $1$ denoting most negative and $20$ denoting most positive control signal).
We separately show the variations of mean positive (negative) sentiment value over the different levels when starting sentiment was negative (positive).
We can observe from Figure~\ref{fig:labelvariation}(a) that the sentiment scores vary over a large range ($0.17$) between the first and last levels for \our{} in case of Yelp when the original sentence is positive; however, \textbf{entangleGen} can achieve a greater negative sentiment score for levels $1-4$. In comparison to these methods, \textbf{ctrlGen} shows a very small drift ($0.1$) in sentiment scores over all the levels with almost zero variation for levels $13-20$.   
This implies that \textbf{ctrlGen} has lesser control in fine tuning sentiment scores over the levels compared to \our{}. Interestingly, if we check the corresponding variation of \emph{Jaccard overlap} in Figure~\ref{fig:labelvariation}(e), we find that the sentences generated by \textbf{ctrlGen} while fine tuning sentiment performs poorly in terms of preserving content. 
For \textbf{entangleGen}, the \emph{Jaccard overlap} is high only for the levels $6-8$. 
On the other-hand, the \emph{Jaccard overlap} scores are high for \our{} over all the levels demonstrating its effectiveness in content preservation while fine tuning sentiment. 

In case of Amazon data, Figure~\ref{fig:labelvariation}(b) shows that we can easily achieve a smooth variation in the negative sentiment scores for \our{} with a reasonably high drift of $0.08$ between the first and last levels. Compared to \our{}, \textbf{ctrlGen} shows negligible drift for levels $11-20$. Figure~\ref{fig:labelvariation}(b) also shows that \textbf{entangleGen} performs poorly in fine tuning the sentiment for Amazon. Looking into the corresponding \emph{Jaccard overlap} scores in Figure~\ref{fig:labelvariation}(f), it can be easily observed that \our{} achieves higher \emph{Jaccard overlap} scores compared to other methods with low scores only for the last few levels. Similar observation holds true for the opposite case when the original sentence is negative across all datasets as evident from Figures~\ref{fig:labelvariation}(c)-{(d) and Figures~\ref{fig:labelvariation}(g)-(h). This signifies the fact that \our{} can achieve a superior fine tuning of sentiment while preserving the content of generated sentences compared to the state-of-the-art methods. Since we could not observe any significant variation in sentiment scores of generated sentences for IMDB over different levels for any of the  methods, we have not discussed these results for the IMDB dataset. 

Further, we have sampled an original sentence from the data and have shown the corresponding generated sentences using \our{} for different levels in Table~\ref{tab:example}. 
As evident from this table, we are able to preserve the content and semantics of the original sentence while still observing a smooth interpolation in the degree of sentiment transfer.
An interesting observation is that though the exact word overlap is low for the generated sentences with very high sentiment scores (e.g., $l_1$ and $l_{20}$ in Table~\ref{tab:example}), they are semantically similar to the original sentence. We show more qualitative results in Appendix B. 





\begin{table}[]
	\small
	\scalebox{0.85}{
	\begin{tabular}{l|l|l}
		\hline
		levels& Original Sentence - Positive                                             & Original Sentence - Negative                           \\ \hline
		original & i ' m very impressed with the level of \textcolor{blue}{care} here !              & but their \textcolor{blue}{inventory} was questionable !      \\ \hline
		$l_1$   & i am \textbf{totally disappointed} by the help  \textbf{completely not} \textcolor{blue}{caring} .         & 
		but their selection was \textbf{overpriced} !
		\\ \hline
			$l_5$   & i am \textbf{totally disappointed} with the compassion ( \textbf{not} \textcolor{blue}{caring} ) .         & but their was \textcolor{blue}{inventory} was \textbf{ridiculous} ! \\ \hline
		$l_{10}$    & i am \textbf{totally disappointed} by the owners ... \textbf{not} \textcolor{blue}{caring} . & but their \textcolor{blue}{inventory} was \textbf{awful} !         \\ \hline
		$l_{13}$    & i am \textbf{not totally disappointed} by the \textcolor{blue}{caring} staff !             & but their \textcolor{blue}{inventory} was \textbf{empty} !             \\ \hline
			$l_{16}$   & i am \textbf{highly recommend}  - this resort \textbf{truly\textcolor{blue}{ care}} !   & their \textcolor{blue}{inventory} was \textbf{ impressive} !            \\ \hline
			$l_{20}$ & i am \textbf{truly kind of `impressed' }- oh \textbf{highly recommend} this staff ! & their selection was \textbf{outstanding} !        \\ \hline
	\end{tabular}}
\caption{Fine grained sentiment control of sentences from extreme negative to extreme positive. \textbf{Bold} letters indicates words with sentiment and the color \textcolor{blue}{blue } indicates important words to be preserved.}
	\label{tab:example}
\end{table}
\subsection{Ablation study}
\todo{I would suggest changing the name of the section -- Expressivenes of $z_a$ or something like that}
Here we perform an ablation study by demonstrating the importance of the last dimension $\zb_a$ of the representation $\zb_f$ in capturing sentiment. As we ensure independence of every dimension, we calculate the correlation of every dimension of $\zb_f$ with the sentiment labels in the test data. We observe that $\zb_a$ achieves the highest correlation of $0.72$ in Yelp and $0.42$ in Amazon. We further train a logistic regression classifier with $\zb_a$ of training data as a feature to predict sentiment labels, and we achieve a high accuracy of $0.85$ and $0.64$ on test data in Yelp and Amazon respectively. While training with the most correlated dimension of $\zb_f$ other than $\zb_a$, with a correlation of $0.12$ for Yelp and $0.14$ for Amazon, we achieve an accuracy of only $0.52$ and $0.58$ respectively. This implies that $\zb_a$ is the most expressive dimension for capturing sentiment in comparison to any other dimension.

\section{Conclusion}
The major contribution of this paper is to propose \our{} which 
consists of a carefully designed hierarchical architecture to maintain both the disentangled feature representation and entangled sentence representation.
The invertible normalizing flow as a transformation module between the two representation layers of \our{} 
enables learning of complex interdependency between
disentangled feature and entangled sentence representation without the loss of information. 
Such a design choice is key to achieving accurate fine tuning of sentiment while keeping the content intact. This is
a key achievement considering the difficulty of the problem and modest performance of state-of-the-art techniques. 
Extensive experiments on real-world datasets emphatically establish the well-rounded performance of \our{} and its
superiority over the baselines. 
\bibliography{main}

\begin{thebibliography}{10}

\bibitem{bahdanau2014neural}
D.~Bahdanau, K.~Cho, and Y.~Bengio.
\newblock Neural machine translation by jointly learning to align and
  translate.
\newblock {\em arXiv preprint arXiv:1409.0473}, 2014.

\bibitem{bowman2015generating}
S.~R. Bowman, L.~Vilnis, O.~Vinyals, and Dai.
\newblock Generating sentences from a continuous space.
\newblock {\em arXiv preprint arXiv:1511.06349}, 2015.

\bibitem{chen2016infogan}
X.~Chen, Y.~Duan, R.~Houthooft, J.~Schulman, I.~Sutskever, and P.~Abbeel.
\newblock Infogan: Interpretable representation learning by information
  maximizing generative adversarial nets.
\newblock In {\em Advances in neural information processing systems}, pages
  2172--2180, 2016.

\bibitem{devlin2018bert}
J.~Devlin, M.-W. Chang, K.~Lee, and K.~Toutanova.
\newblock Bert: Pre-training of deep bidirectional transformers for language
  understanding.
\newblock {\em arXiv preprint arXiv:1810.04805}, 2018.

\bibitem{diao2014jointly}
Q.~Diao, M.~Qiu, C.-Y. Wu, A.~J. Smola, J.~Jiang, and C.~Wang.
\newblock Jointly modeling aspects, ratings and sentiments for movie
  recommendation (jmars).
\newblock In {\em Proceedings of the 20th ACM SIGKDD international conference
  on Knowledge discovery and data mining}, pages 193--202, 2014.

\bibitem{dinh2016density}
L.~Dinh, J.~Sohl-Dickstein, and S.~Bengio.
\newblock Density estimation using real nvp.
\newblock {\em arXiv preprint arXiv:1605.08803}, 2016.

\bibitem{fu2018style}
Z.~Fu, X.~Tan, N.~Peng, D.~Zhao, and R.~Yan.
\newblock Style transfer in text: Exploration and evaluation.
\newblock In {\em Thirty-Second AAAI Conference on Artificial Intelligence},
  2018.

\bibitem{higgins2017beta}
I.~Higgins, L.~Matthey, A.~Pal, C.~Burgess, X.~Glorot, M.~Botvinick,
  S.~Mohamed, and A.~Lerchner.
\newblock beta-vae: Learning basic visual concepts with a constrained
  variational framework.
\newblock {\em ICLR}, 2(5):6, 2017.

\bibitem{hu2017toward}
Z.~Hu, Z.~Yang, X.~Liang, R.~Salakhutdinov, and E.~P. Xing.
\newblock Toward controlled generation of text.
\newblock In {\em Proceedings of the 34th International Conference on Machine
  Learning-Volume 70}, pages 1587--1596. JMLR. org, 2017.

\bibitem{john2018disentangled}
V.~John, L.~Mou, H.~Bahuleyan, and O.~Vechtomova.
\newblock Disentangled representation learning for non-parallel text style
  transfer.
\newblock {\em arXiv preprint arXiv:1808.04339}, 2018.

\bibitem{kannan2017adversarial}
A.~Kannan and O.~Vinyals.
\newblock Adversarial evaluation of dialogue models.
\newblock {\em arXiv preprint arXiv:1701.08198}, 2017.

\bibitem{kim2018disentangling}
H.~Kim and A.~Mnih.
\newblock Disentangling by factorising.
\newblock {\em arXiv preprint arXiv:1802.05983}, 2018.

\bibitem{kingma2016improved}
D.~P. Kingma, T.~Salimans, R.~Jozefowicz, X.~Chen, I.~Sutskever, and
  M.~Welling.
\newblock Improved variational inference with inverse autoregressive flow.
\newblock In {\em Advances in neural information processing systems}, pages
  4743--4751, 2016.

\bibitem{kumar2017variational}
A.~Kumar, P.~Sattigeri, and A.~Balakrishnan.
\newblock Variational inference of disentangled latent concepts from unlabeled
  observations.
\newblock {\em arXiv preprint arXiv:1711.00848}, 2017.

\bibitem{li2017deep}
P.~Li, W.~Lam, L.~Bing, and Z.~Wang.
\newblock Deep recurrent generative decoder for abstractive text summarization.
\newblock {\em arXiv preprint arXiv:1708.00625}, 2017.

\bibitem{logeswaran2018content}
L.~Logeswaran, H.~Lee, and S.~Bengio.
\newblock Content preserving text generation with attribute controls.
\newblock In {\em Advances in Neural Information Processing Systems}, pages
  5103--5113, 2018.

\bibitem{rezende2015variational}
D.~J. Rezende and S.~Mohamed.
\newblock Variational inference with normalizing flows.
\newblock {\em arXiv preprint arXiv:1505.05770}, 2015.

\bibitem{shen2017style}
T.~Shen, T.~Lei, R.~Barzilay, and T.~Jaakkola.
\newblock Style transfer from non-parallel text by cross-alignment.
\newblock In {\em Advances in neural information processing systems}, pages
  6830--6841, 2017.

\bibitem{singh2018sentiment}
A.~Singh and R.~Palod.
\newblock Sentiment transfer using seq2seq adversarial autoencoders.
\newblock {\em arXiv preprint arXiv:1804.04003}, 2018.

\bibitem{socher2013recursive}
R.~Socher, A.~Perelygin, J.~Wu, J.~Chuang, C.~D. Manning, A.~Ng, and C.~Potts.
\newblock Recursive deep models for semantic compositionality over a sentiment
  treebank.
\newblock In {\em Proceedings of the 2013 conference on empirical methods in
  natural language processing}, pages 1631--1642, 2013.

\bibitem{sutskever2014sequence}
I.~Sutskever, O.~Vinyals, and Q.~V. Le.
\newblock Sequence to sequence learning with neural networks.
\newblock In {\em Advances in neural information processing systems}, pages
  3104--3112, 2014.

\bibitem{turc2019}
I.~Turc, M.-W. Chang, K.~Lee, and K.~Toutanova.
\newblock Well-read students learn better: On the importance of pre-training
  compact models.
\newblock {\em arXiv preprint arXiv:1908.08962v2}, 2019.

\bibitem{tustison2009introducing}
N.~Tustison and J.~Gee.
\newblock Introducing dice, jaccard, and other label overlap measures to itk.
\newblock {\em Insight J}, 2, 2009.

\bibitem{vaswani2017attention}
A.~Vaswani, N.~Shazeer, N.~Parmar, J.~Uszkoreit, L.~Jones, A.~N. Gomez,
  {\L}.~Kaiser, and I.~Polosukhin.
\newblock Attention is all you need.
\newblock In {\em Advances in neural information processing systems}, pages
  5998--6008, 2017.

\bibitem{wang2019controllable}
K.~Wang, H.~Hua, and X.~Wan.
\newblock Controllable unsupervised text attribute transfer via editing
  entangled latent representation.
\newblock In {\em Advances in Neural Information Processing Systems}, pages
  11034--11044, 2019.

\bibitem{zhang2018shaped}
Y.~Zhang, N.~Ding, and R.~Soricut.
\newblock Shaped: Shared-private encoder-decoder for text style adaptation.
\newblock {\em arXiv preprint arXiv:1804.04093}, 2018.

\bibitem{zhang2018style}
Z.~Zhang, S.~Ren, S.~Liu, J.~Wang, P.~Chen, M.~Li, M.~Zhou, and E.~Chen.
\newblock Style transfer as unsupervised machine translation.
\newblock {\em arXiv preprint arXiv:1808.07894}, 2018.

\bibitem{zhao2017adversarially}
J.~Zhao, Y.~Kim, K.~Zhang, A.~M. Rush, and Y.~LeCun.
\newblock Adversarially regularized autoencoders.
\newblock {\em arXiv preprint arXiv:1706.04223}, 2017.

\end{thebibliography}
\bibliographystyle{abbrv}

\clearpage
\appendix
\section{Parameter Setting}
The sentence encoder is designed using pre-trained BERT-base-uncased model (embedding dim = $768$) followed by 2-layer feed-forward network with hidden dim $200$. The output of the same is the sentence embedding which is of dimension
$256$ for Yelp and $300$ for Amazon, and IMDB. The flow network is designed as R-NVP with $T = 3$ and each $\psi_t$ is designed as three layer  feed forward network with $tanh$ activation function for the initial two layers and hidden dimension is $100$ for the intermediate layers. The scaling network for sentiment classification is designed as a two dimensional vector $[-1, 1]$. The sentence decoder is designed as a gated recurrent unit where output of each step is passed through a fully connected feed-forward network to convert it to a logit of length of the vocabulary size. To avoid vanishingly small KL term in the VAE module Eq. 11, we use a KL term weight linearly annealing from 0 to 1 during training. The weighing parameters $\beta$ and $\gamma$ are set to $10$ for feature supervision and disentanglement .

\section{Qualitative samples}
In Table ~\ref{tab:comparative}, we have provided some comparative examples of the sentiment fine-tuning results. It can be seen that \textbf{ctrlGen} provides both positive and negative polarity sentences, but there is very less content overlap in comparison to \textbf{entangleGen} and \textbf{DE-VAE}. On further scrutiny we discover, it can only retain content for smaller length sentences with less content words. Some examples are given in Table ~\ref{tab:short}.
\begin{table}[ht]
	\small
	\begin{tabular}{l|l|l|l}
		\hline
		&\textbf{ DE-VAE} & \textbf{entangleGen} & \textbf{ctrlGen} \\ \hline
		Orig.& \multicolumn{3}{c}{the house fried rice and egg rolls are the best in phoenix .} \\ \hline
	$l_1$	& \begin{tabular}[c]{@{}l@{}}the fried rice dishes are not  acceptable \\,and extremely disappointed .\end{tabular} &  \begin{tabular}[c]{@{}l@{}} the house too way \\and egg should ? ''\end{tabular}  & rustrated ! \\ \hline
$l_5$		& \begin{tabular}[c]{@{}l@{}}the eggs benedict fried rice and\\  the rice dishes are not acceptable .\end{tabular} & \begin{tabular}[c]{@{}l@{}}the house too way and egg rolls \\ were the barely in nothing ?\end{tabular} & this is n't the only restaurant . \\ \hline
$l_{8}$		& \begin{tabular}[c]{@{}l@{}}the rice dishes of fried rice , \\ which is not the perfect\end{tabular} & \begin{tabular}[c]{@{}l@{}}the house fried rice and egg rolls \\ were the only in phoenix table .\end{tabular} & their fish is soggy . \\ \hline
$l_{10}$			& \begin{tabular}[c]{@{}l@{}}the rice dishes up fried rice , \\ which is not the wonderful .\end{tabular} & \begin{tabular}[c]{@{}l@{}}the house fried rice and egg rolls \\ are the best in phoenix only .\end{tabular} & i will return to give you right new . \\ \hline
	$l_{12}$		& \begin{tabular}[c]{@{}l@{}}the rice combo is the average , \\ eggs and authentic.\end{tabular} & - & the grand staff is nice . \\ \hline
	$l_{16}$		& \begin{tabular}[c]{@{}l@{}}the rice ( the chicken ) are the best ,\\  and the same for the staff\end{tabular} & - & friendly staff is impeccably neat \\ \hline
	$l_{20}$		& the rice ( the chicken ) are the best & - & bar tender and professional . \\ \hline
	\end{tabular}
\caption{Comparative example with DE-VAE and other methods}
\label{tab:comparative}
\end{table}

\begin{table}[ht]
	\small
	\centering
	\begin{tabular}{l|l|l}
		\hline
		&  \textbf{DE-VAE}&\textbf{ctrlGen}  \\ \hline
		Orig.& \multicolumn{2}{l}{food was good , but service was great .} \\ \hline
	 $l_1$	& aweful service & hate the food , service just . \\ \hline
	 $l_{20}$	& food was great service was good too . & great food . \\ \hline
		\hline
	 Orig.	& \multicolumn{2}{l}{will make this place a regular staple .} \\ \hline
	 $l_1$	& will make this place aweful & do n't make this place . \\ \hline
	$l_{20}$	& will make this place great & i love this place . \\ \hline
	\end{tabular}
\caption{Sample generated text from shorter original text}
\label{tab:short}
\end{table}
\begin{figure}[ht]
	\subfloat[Training time]{\includegraphics[height=0.34\textwidth,keepaspectratio]{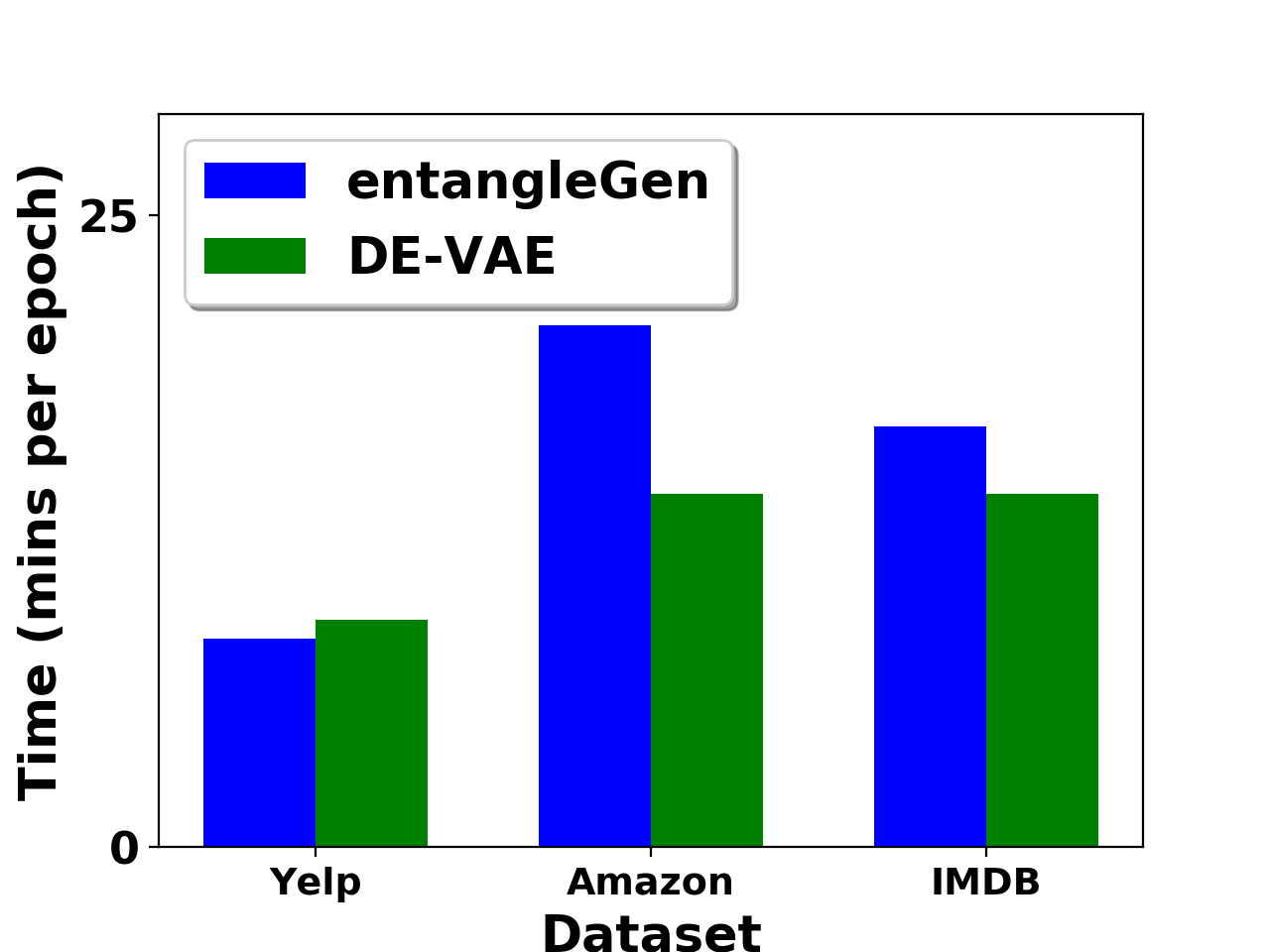}}
	\subfloat[Generation time]{\includegraphics[height=0.34\textwidth,keepaspectratio]{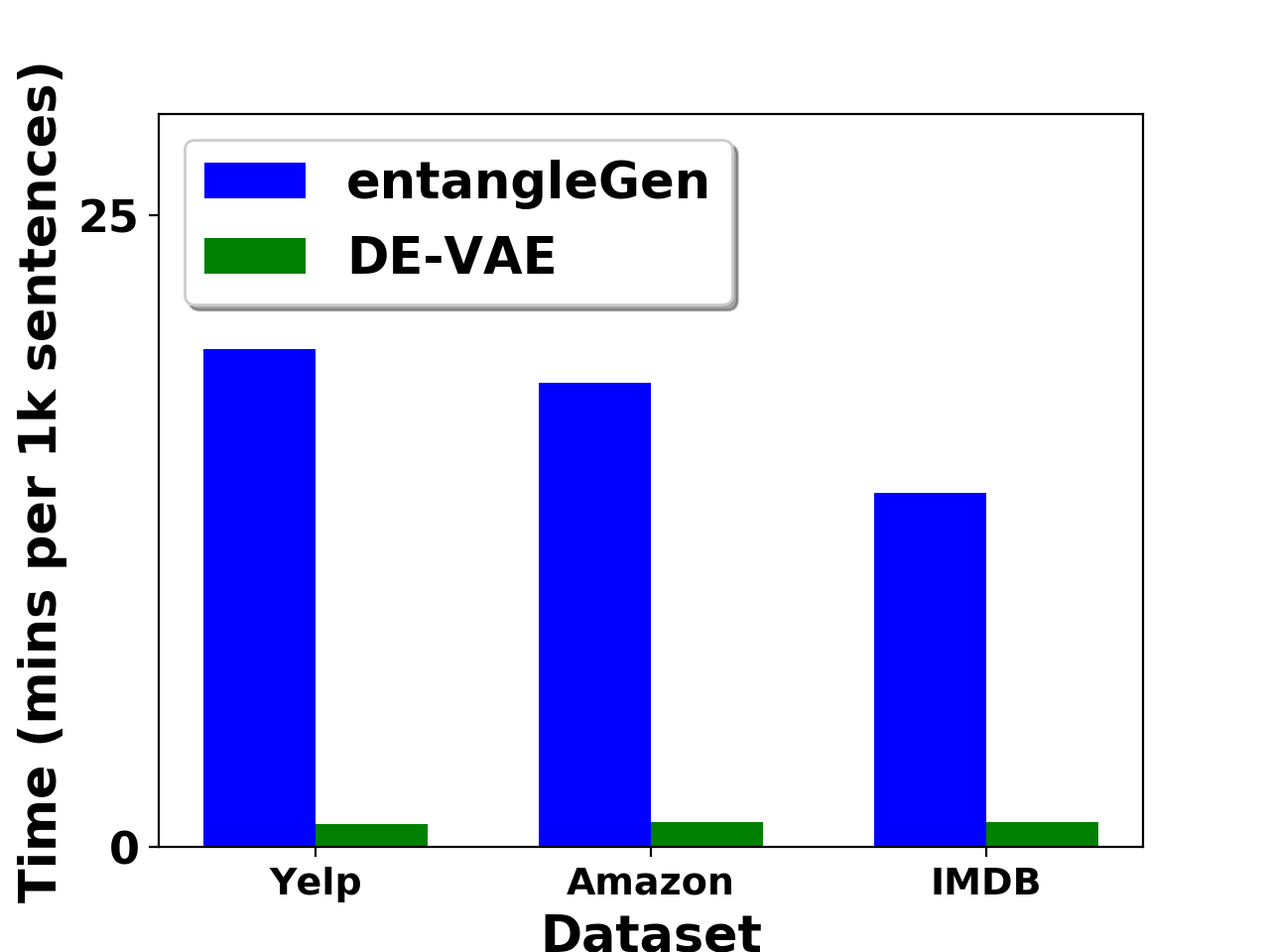}}
	\caption{(a) the time taken (per epoch) for training by  DE-VAE and entangleGen on different datasets. (b) the time taken to generate $1$K sentences by DE-VAE and entangleGen on different datasets.}
	\label{fig:time_compare}
\end{figure}
\section{Training time comparison}

In this section we provide a comparative analysis of training time and sampling time of \textbf{DE-VAE} with \textbf{entangleGen}. Fig ~\ref{fig:time_compare} shows that \textbf{DE-VAE} is much faster than that of \textbf{entangleGen} for both cases.

\end{document}